\documentclass{article}

\usepackage[T1]{fontenc}
\usepackage[utf8]{inputenc}
\usepackage{color}
\usepackage[brazilian,english]{babel}

\usepackage{amsmath}
\usepackage{amssymb}
\usepackage{amsfonts}

\usepackage{graphicx}
\usepackage{psfrag}
\usepackage{subcaption}

\usepackage{amsthm}
\usepackage{thmtools}
\usepackage{thm-restate}

\usepackage{tikz}
\usepackage{pgfplots}
\usepgfplotslibrary{patchplots}
\usetikzlibrary{fit,positioning}

\usepackage{algpseudocode}

\usepackage{xspace}
\usepackage{multirow}
\usepackage{mathtools}
\usepackage{cancel}

\usepackage{nips15submit_e}
\usepackage{times}
\usepackage{hyperref}
\usepackage{url}

\newcommand{\R}{\mathbb{R}\xspace}
\newcommand{\X}{\mathcal{X}\xspace}

\title{Multi-Objective Optimization for Self-Adjusting Weighted Gradient in
Machine Learning Tasks}

\author{
Conrado S. Miranda\\
Sch. of Electrical and Computer Engineering\\
University of Campinas\\
Campinas, Brazil \\
\texttt{conrado@dca.fee.unicamp.br} \\
\And
Fernando J. Von Zuben \\
Sch. of Electrical and Computer Engineering\\
University of Campinas\\
Campinas, Brazil \\
\texttt{vonzuben@dca.fee.unicamp.br} \\
}

\nipsfinalcopy

\begin{document}

\maketitle

\begin{abstract}
  Much of the focus in machine learning research is placed in creating new
  architectures and optimization methods, but the overall loss function is
  seldom questioned. This paper interprets machine learning from a
  multi-objective optimization perspective, showing the limitations of the
  default linear combination of loss functions over a data set and introducing
  the hypervolume indicator as an alternative. It is shown that the gradient of
  the hypervolume is defined by a self-adjusting weighted mean of the individual
  loss gradients, making it similar to the gradient of a weighted mean loss but
  without requiring the weights to be defined a priori. This enables an inner
  boosting-like behavior, where the current model is used to automatically place
  higher weights on samples with higher losses but without requiring the use of
  multiple models. Results on a denoising autoencoder show that the new
  formulation is able to achieve better mean loss than the direct optimization
  of the mean loss, providing evidence to the conjecture that self-adjusting the
  weights creates a smoother loss surface.
\end{abstract}

\section{Introduction}
\label{sec:introduction}
Many machine learning algorithms, including neural networks, can be divided into
three parts: the model, which is used to describe or approximate the structure
present in the training data set; the loss function, which defines how well an
instance of the model fits the samples; and the optimization method, which
adjusts the model's parameters to improve the error expressed by the loss
function. Obviously these three parts are related, and the generalization
capability of the obtained solution depends on the individual merit of each one
of the three parts, and also on their interplay.

Most of current research in machine learning focuses on creating new
models~\cite{koller2009probabilistic,bengio2009learning}, for the different
applications and data types, and new optimization
methods~\cite{bennett2006interplay}, which may allow faster convergence, more
robustness, and a better chance to escape from poor local minima. However, focus
is rarely attributed to the cost functions used in different applications.

Many cost functions come from statistical models~\cite{bishop2006pattern}, such
as the quadratic error or cross-entropy. These cost functions, in their linear
formulations, usually define convex loss
functions~\cite{bishop2006pattern,banerjee2005clustering}. Moreover, when
building the statistical model of a sample set, frequently the total cost
becomes the sum of losses for each sample. Although this methodology is sound,
it can be problematic in real-world applications involving more complex models.

The first problem, discussed more extensively in Sec.~\ref{sec:moo}, is that
linear combination of losses can reduce the number of solutions achievable by
the optimization algorithm. The second problem is that this cost may not reflect
the behavior expected by the user.

Consider, for instance, a classification problem with samples $x_1$ and $x_2$
and two choices of parameters $\theta_1$ and $\theta_2$. For $\theta_1$, the
classifier has confidences $99\%$ and $49\%$ for the correct classes of $x_1$
and $x_2$, which means that $x_2$ is classified incorrectly. For $\theta_2$,
they are classified correctly with confidences $90\%$ and $53\%$, respectively.
Most machine learning users would prefer using $\theta_2$, because $x_1$ has a
high-confidence correct prediction in both cases while $x_2$ is classified
incorrectly if $\theta_1$ is used. Nonetheless, the mean cross-entropy loss,
which is the standard cost for classification problems, places higher cost to
$\theta_2$ than to $\theta_1$, so the optimizer will choose $\theta_1$.

Although this toy example may not be representative of all the trade-offs
involved, it clearly reflects the conflict between what the user wants and what
the cost function expresses. One possible solution to this problem is the use of
boosting techniques to increase the weight of incorrectly classified samples and
training multiple classifiers, as discussed in Sec.~\ref{sec:gradient}, but this
requires multiple models and may not be robust to noise~\cite{long2010random}.
Moreover, this solution still restricts the solutions achievable by the
optimization algorithm due to the linear combination of losses.

This paper presents a solution to this problem by evaluating the model-fitting
problem in a multi-objective optimization perspective, of which the standard
linear combination of losses is a particular case, and uses another method for
transforming the problem into a single-objective optimization, which allows the
same standard optimization algorithms to be used. It will be shown that the
gradient of the new objective automatically provides higher weights for samples
with higher losses, like boosting methods, but uses a single model, with the
current parameters working as the previous model in boosting. It is important to
highlight that, although this paper focuses on optimization by gradient, the
method proposed also works for hill-climbing on discrete problems.

The author's conjecture that automatically placing more learning pressure on
samples with high losses may improve the overall generalization, because it may
remove small bumps in the error surface. Therefore, the multi-objective
formulation allows better minima to be reached. Experimental results presented
in this paper provide evidence for this conjecture.

The paper is organized as follows. Section~\ref{sec:moo} provides an overview of
multi-objective optimization, properly characterizing the maximization of the
hypervolume as a performance indicator for the learning system, given the
data set and the loss function. Section~\ref{sec:gradient} deals with a
gradient-based approach to maximize the hypervolume, and shows that the gradient
operator promotes the self-adjustment of the weights associated with each loss
function. Sections~\ref{sec:experiment} and~\ref{sec:results} describe the
experiments performed and the results obtained, respectively. Finally,
Section~\ref{sec:conclusion} provides the concluding remarks and future research
directions.
\section{Multi-objective optimization}
\label{sec:moo}
Multi-objective optimization is a generalization of the traditional
single-objective optimization, where the problem is composed of multiple
objective functions $f_i \colon \X \to \mathcal Y_i,i \in \{1,\ldots,N\}$, where
$\mathcal Y_i \subseteq \R$~\cite{deb2014multi}. Using the standard notation,
the problem can be described by:
\begin{equation*}
  \min_{x \in \X} ~ (f_1(x), f_2(x),\ldots,f_N(x)),
\end{equation*}
where $\X$ is the decision space and includes all constraints of the
optimization.

If some of the objectives have the same minima, then the redundant objectives
can be ignored during optimization. However, if their minima are different, for
example $f_1(x) = x^2$ and $f_2(x) = (x-1)^2$, then there is not a single
optimal point, but a set of different trade-offs between the objectives. A
solution that establishes an optimal trade-off, that is, it is impossible to
reduce one of the objectives without increasing another, is said to be
efficient. The set of efficient solutions is called the Pareto set and its
counterpart in the objective space $\mathcal Y = \mathcal Y_1 \times \cdots
\times \mathcal Y_N$ is called the Pareto frontier.

A common approach in optimization used to deal with multi-objective problems is
to combine the objectives linearly~\cite{deb2014multi,boyd2004convex}, so that
the problem becomes
\begin{equation}
  \label{eq:linear_combination}
  \min_{x \in \X} ~ \sum_{i=1}^N w_i f_i(x),
\end{equation}
where the weight $w_i \in \R^+$ represents the importance given to objective
$i,i \in \{1,\ldots,N\}$.

Although the optimal solution of the linearly combined problem is guaranteed to
be efficient, it is only possible to achieve any efficient solution when the
Pareto frontier is convex~\cite{boyd2004convex}. This means that more desired
solutions may not be achievable by performing a linear combination of the
objectives.

As an illustrative example, consider the problem where $\X = [0,1]$ and the
Pareto frontier is almost linear, with $f_1(x) \approx x$ and $f_2(x) \approx
1-x$, but slightly concave. Any combination of weights $w_1$ and $w_2$ will only
be able to provide $x=0$ or $x=1$, depending on the specific weights, but a
solution close to 0.5 might be more aligned with the trade-offs expected by the
user.

\subsection{Hypervolume indicator}
\label{sec:moo:hypervolume}
Since the linear combination of objectives is not going to work properly on
non-convex Pareto frontiers, it is desirable to investigate other forms of
transforming the multi-objective problem into a single-objective one, which
allows the standard optimization tools to be used.

One common approach in the multi-objective literature is to resort to the
hypervolume indicator~\cite{zitzler2007hypervolume}, which is frequently used to
analyze a set of candidate solutions~\cite{zitzler2003performance} and can be
expensive in such cases~\cite{beume2009complexity}. However for a single
solution, its logarithm can be written as:
\begin{equation}
  \label{eq:hypervolume}
  \log H_z(x) = \sum_{i=1}^N \log(z_i - f_i(x)),
\end{equation}
where $z \in \R^N$ is called the Nadir point and $z_i > f_i(x), \forall x \in
\X$. The problem then becomes maximizing the hypervolume over the domain, and
this optimization is able to achieve a larger number of efficient points,
without requiring convexity of the Pareto frontier~\cite{auger2009theory}.

Among the many properties of the hypervolume, two must be highlighted in this
paper. The first is that the hypervolume is monotonic in the objectives, which
means that any improvement in any objective causes the hypervolume to increase,
which in turn is aligned with the loss minimization. The maximum of the
single-solution hypervolume is a point in the Pareto set, which means that the
solution is efficient.

The second property is that, like the linear combination, it also maintains some
shape information from the objectives. If the objectives are convex, then their
linear combination is convex and the hypervolume is concave, since $-f_i(x)$ is
concave and the logarithm of a concave function is concave.

Therefore, the hypervolume has the same number of parameters as the linear
combination, one for each objective, and also maintains optimality, but may be
able to achieve solutions more aligned with the user's expectations.

Consider the previous example with $f_1(x) \approx x$ and $f_2(x) \approx 1-x$.
If $z_1 = z_2$, which indicates that both objectives are equally important a
priori, then the hypervolume maximization will indeed find a solution close to
$x = 0.5$. So the parameter $z$ defines prior inverse importances that are
closer to what one would expect from weights: if they are equal, the solver
should try to provide a balanced relevance to the objectives.

Therefore, the hypervolume maximization with a single model tries to find a good
solution to all objectives at the same time, achieving the best compromise
possible given the prior inverse importances $z$. Note that, contrary to the
weights $w$ in the linear combination, the values $z$ actually reflect the
user's intent for the solution found in the optimization.

\subsection{Loss minimization}
A standard objective in machine learning is minimization of some loss function
$l \colon \X \times \Theta \to \R$ over a given data set $X = \{x_1,\ldots,x_N\}
\subset \X$. Note that this notation includes both supervised and unsupervised
learning, as the space $\X$ can include both the samples and their targets.

Using the multi-objective notation, the loss minimization problem is described
as:
\begin{equation*}
  \min_{\theta \in \Theta} ~ (l(x_1,\theta), \ldots, l(x_N,\theta)).
\end{equation*}
Just like in other areas of optimization, the usual approach to solve these
problems in machine learning is the use of linear combination of the objectives,
as shown in Eq.~\eqref{eq:linear_combination}. Common examples of this method
are using the mean loss as the objective to be minimized and defining a
regularization weight~\cite{bishop2006pattern}, where the regularization
characterizes other objectives.

However, as discussed in Sec.~\ref{sec:moo}, this approach limits the number of
solutions that can be obtained, which motivates the use of the hypervolume
indicator. Since the objectives differ only in the samples used for the loss
function and considering that all samples have equal importance\footnote{This is
the same motivation for using the uniform mean loss. If prior importance is
available, it can be used to define the value of $z$, like it would be used to
define $w$ in the weighted mean loss.}, the Nadir point $z$ can have the same
value for all objectives so that the solution found is balanced in the losses.
This value is given by the parameter $\mu$, so that $z_i = \mu, \forall i$. Then
the problem becomes
\begin{equation*}
  \max_{\theta \in \Theta} \log H_\mu(\theta),
\end{equation*}
where
\begin{equation}
  \label{eq:hypervolume_mu}
  \log H_\mu(\theta) = \sum_{i=1}^N \log(\mu - l(x_i,\theta)),
\end{equation}
which, as stated in Sec.~\ref{sec:moo:hypervolume}, maintains the convexity
information of the loss function but has a single parameter $\mu$ to be chosen
along the iterative steps of the optimization.

As will be shown in Sec.~\ref{sec:gradient}, smaller values of $\mu$ place
learning pressure on samples with large losses, improving the worst-case
scenario, while large values of $\mu$ approximates learning to the uniform mean
loss. And, of course, as learning progresses, the relative hardness of each
sample may vary significantly.
\section{Gradient as the operator for self-adjusting the weights}
\label{sec:gradient}
As exemplified in Sec.~\ref{sec:introduction}, the standard metric of average
error can be problematic by providing excessive confidence to some samples while
preventing less-confident samples to improve. A solution to this problem is to
use different weights, as in Eq.~\eqref{eq:linear_combination}, with higher
weights for harder samples instead of using the same weight for all samples.

However, one may not know a priori which samples are harder, which prevents the
straightforward use of the weighted mean. One classical solution to this problem
in classification is the use of boosting techniques~\cite{schapire1990strength},
where a set of classifiers are learnt instead of a single one and they are
merged into an ensemble. When training a new classifier, the errors of the
previous ones on each sample are taken into account to increase the weights of
incorrectly labeled samples, placing more pressure on the new classifier to
predict these samples' labels correctly.

The requirement of multiple models, to form the boosting ensemble, can be hard
to train with very large models and may be too slow to be used in real problems,
since multiple models have to be evaluated. Therefore, a single model that is
able to self-adjust the weights so that higher weights are going to be assigned
to currently harder samples is desired.

Taking the gradient of the losses hypervolume's logarithm, shown in
Eq.~\eqref{eq:hypervolume_mu}, one finds
\begin{equation}
  \frac{\partial \log H_\mu}{\partial \theta} =
  - \sum_{i=1}^N \frac{1}{\mu - l(x_i,\theta)}
  \frac{\partial l(x_i,\theta)}{\partial \theta}
  =
  - \sum_{i=1}^N w_i
  \frac{\partial l(x_i,\theta)}{\partial \theta}, \quad
  w_i = \frac{1}{\mu - l(x_i,\theta)},
\end{equation}
which is similar to a weighted mean loss cost with weights $w_i,i \in
\{1,\ldots,N\},$ replacing the minimization of the loss by the maximization of
the hypervolume. However, the weights $w$ are determined automatically as a
function of $\mu$, so that they do not have to be defined a priori and can
change during the learning algorithm's execution. Moreover, higher cost implies
higher weight, that is,
\begin{equation}
  l(x_i,\theta) > l(x_j,\theta) \Rightarrow w_i > w_j, \quad
  i,j \in \{1,\ldots,N\}
\end{equation}
which is expected, as more pressure must be placed on samples with higher
losses.

Therefore, hypervolume maximization using gradient adjusts the parameters in the
same direction as weighted mean loss minimization with appropriate weights.
Furthermore, as $\mu$ approaches $\max_i l(x_i,\theta)$, the worst case becomes
the main influence in the gradient and the problem becomes similar to minimizing
the maximum loss, while as $\mu$ approaches infinity, the problem becomes
similar to minimizing the uniform mean loss. Thus a single parameter is able to
control the problem's placement between these two extremes.

Moreover, the hypervolume's logarithm, show in Eq.~\eqref{eq:hypervolume}, is
just a sum of the different objectives, like the linear combination shown in
Eq.~\eqref{eq:linear_combination}. This allows the same optimization methods to
be used for maximizing the hypervolume, which includes methods that deal with
large data sets and scalability issues, such as the stochastic gradient descent
with mini-batches.
\section{Experiment}
\label{sec:experiment}
To evaluate the performance of maximizing the hypervolume, experiments were
carried out using a denoising autoencoder~\cite{vincent2008extracting}, which is
a neural network with a hidden layer that tries to reconstruct the input at the
output, given a corrupted version of the input. The purpose is to reconstruct
the digits in the MNIST data set, with 50000 training samples and 10000
validation and test samples each, as usual with this data set. Since the uniform
mean loss is the usual objective for this kind of problem, it is used as a
baseline, establishing a reference for comparison with the new proposed
formulation, and as the indicator for comparison between the baseline and the
hypervolume method.

In order to be able to use the same learning rate for the hypervolume and mean
loss problems, and since the mean loss already has normalized weights, the
parameter's adjustment step was divided by the sum of the weights $w_i, i \in
\{1,\ldots,N\},$ computed using the hypervolume gradient in order to normalize
the weights.

The hypervolume maximization depends on the definition of the parameter $\mu$,
as discussed in Sec.~\ref{sec:gradient}. Since it must be larger than the
objectives when the gradient is computed, the parameter $\mu$ used in the
experiments is defined by:
\begin{equation}
  \mu^{(t)} = \max_i l(x_i, \theta^{(t)}) + \epsilon^{(t)}, \quad
  \epsilon^{(t)} = \epsilon^{(0)} + \kappa t, \quad
  \epsilon^{(0)}, \kappa \ge 0,
\end{equation}
where $t$ is the epoch number, $\theta^{(t)}$ are the parameters at epoch $t$,
$\epsilon^{(0)}$ is the initial slack constant, and $\kappa$ is a user-chosen
constant that can decrease the learning pressure over bad examples, as the
number of epochs increase. For the stochastic gradient using mini-batches,
$\theta^{(t)}$ corresponds to the parameters at that point, but $\epsilon^{(t)}$
is only increased after the full epoch is completed.

This way of defining the Nadir point removes the requirement of choosing an
absolutely good value by allowing a value relative to the current losses to be
determined, making its choice easier. The parameters were arbitrarily chosen as
$\epsilon^{(0)} = 1$ and $\kappa = 1$ for the experiments, so that the problem
becomes more similar to the mean loss problem as the learning progresses, since
the mean loss is used as the performance indicator.

The optimization was performed using gradient descent with learning rate $0.1$
and mini-batch size of $500$ samples for $100$ epochs, with the cross-entropy as
the loss function. A total of $50$ runs with different initializations of the
random number generator were used.

Salt-and-pepper noise was added to the training images with varying probability
$p$, but with equal probabilities of black or white corruption, and $500$ hidden
units were used in the denoising autoencoder. Notice that the number of inputs,
and consequently the number of outputs, is $784$, since the digits are
represented by $28\times28$ images. In order to evaluate the performance, the
mean and maximum losses over the training, validation, and test sets were
computed without corruption of the inputs.
\section{Results}
\label{sec:results}

\begin{figure}[h]
  \centering
  \begin{subfigure}{0.45\linewidth}
    \psfrag{iterations}[t][c]{\scriptsize Epoch}
    \psfrag{h0-h1}[c][c]{\scriptsize Difference}
    \includegraphics[width=\linewidth]{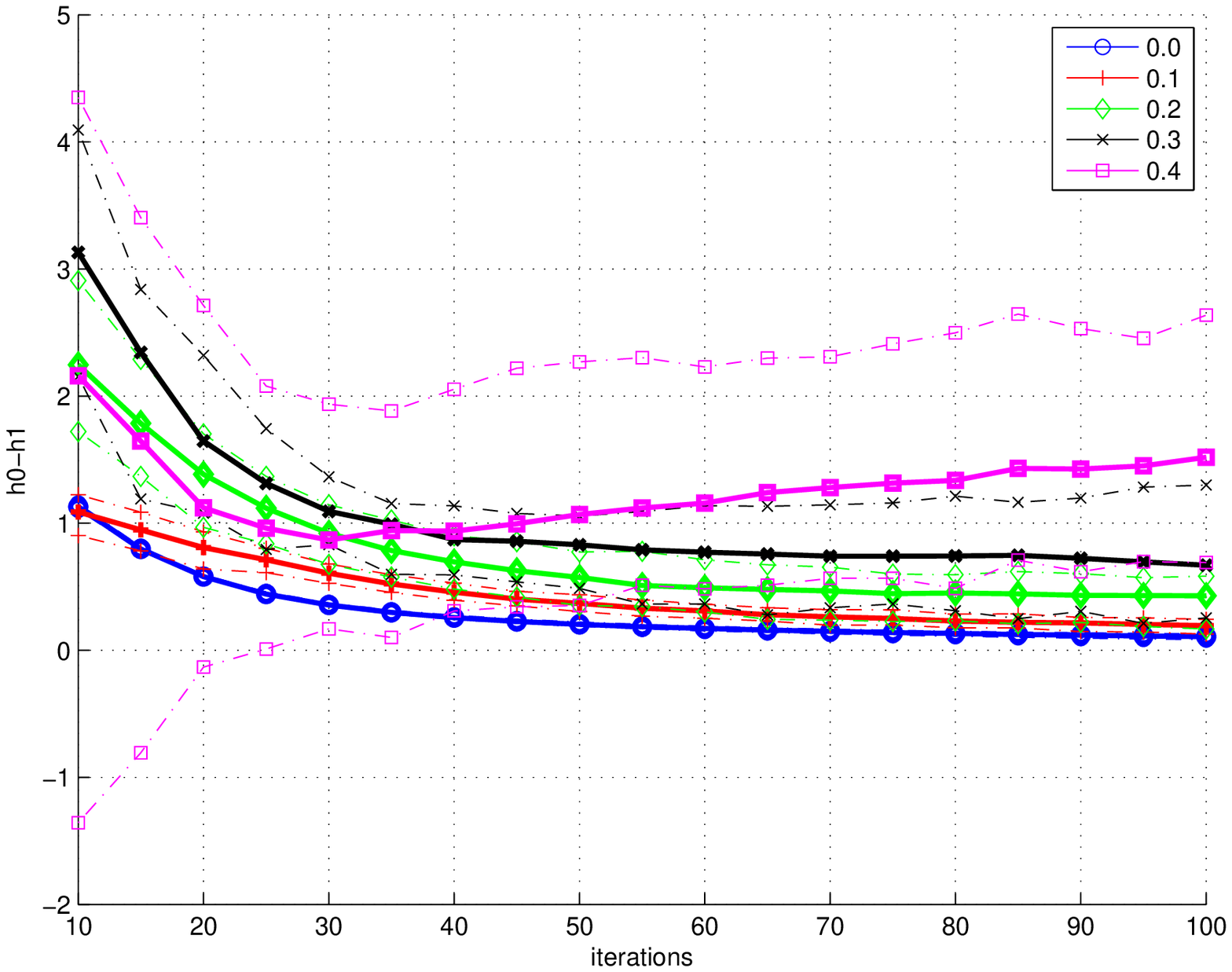}
    \caption{Training}
  \end{subfigure}
  \begin{subfigure}{0.45\linewidth}
    \psfrag{iterations}[t][c]{\scriptsize Epoch}
    \psfrag{h0-h1}[c][c]{\scriptsize Difference}
    \includegraphics[width=\linewidth]{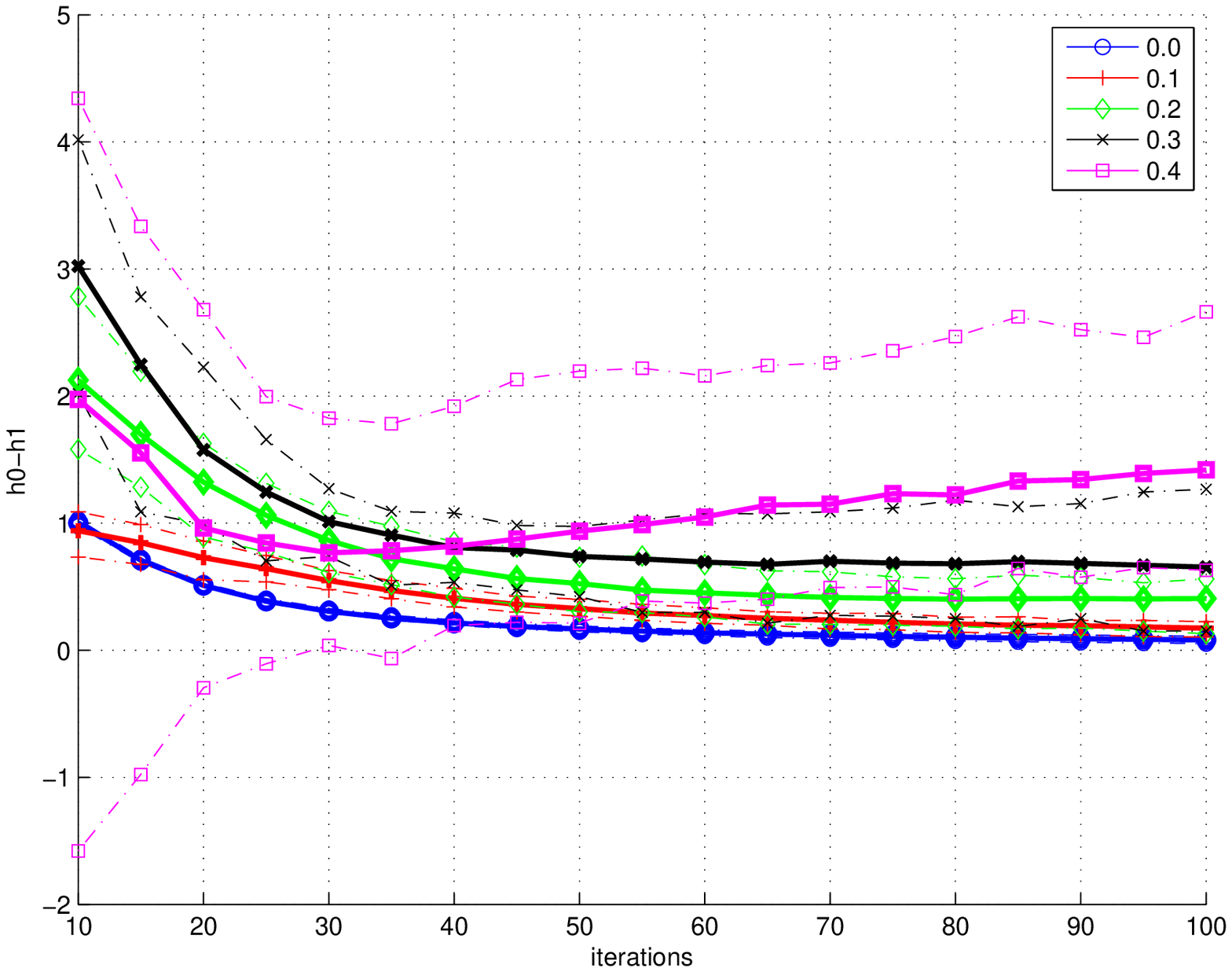}
    \caption{Validation}
  \end{subfigure}
  \\
  \begin{subfigure}{0.45\linewidth}
    \psfrag{iterations}[t][c]{\scriptsize Epoch}
    \psfrag{h0-h1}[c][c]{\scriptsize Difference}
    \includegraphics[width=\linewidth]{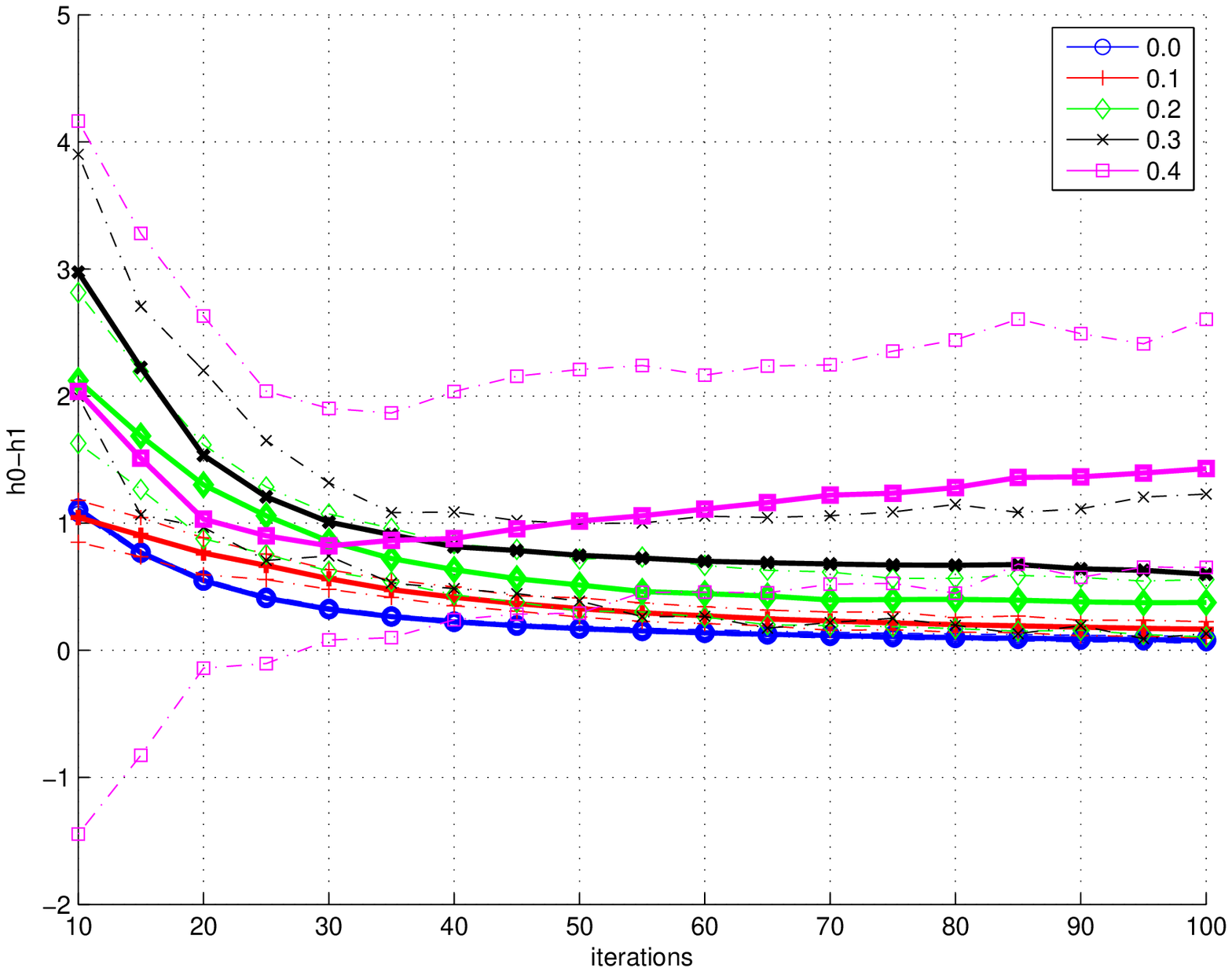}
    \caption{Test}
  \end{subfigure}
  \caption{Difference of the mean loss between mean loss minimization and
    hypervolume maximization (values above zero mean improvement) for different
    corruption probabilities $p$. The median over the runs is shown in solid
  line, while the upper and lower bounds are shown in dashed lines.}
  \label{fig:results_mean}
\end{figure}

Figure~\ref{fig:results_mean} shows the mean loss over the iterations for the
training, validation, and test sets of the MNIST data set for different
corruption probabilities $p$. The plotted values are the difference between the
standard mean loss minimization, used as a baseline, and the mean loss when
maximizing the hypervolume. Positive values mean that the hypervolume
maximization provides lower mean losses, which corresponds to better results.
The same initial conditions and noises were used for the baseline and the
proposed method, so that the difference in performance in a single run is caused
by the difference of objectives and is not caused by different random numbers.

From Fig.~\ref{fig:results_mean}, it is clear that the hypervolume method
provides better results than the baseline for all corruption levels, indicating
a better fit of the parameters. Moreover, the difference becomes larger as the
noise level increases, indicating that the proposed method is more robust and
guides to better generalization.

It is important to highlight that the hypervolume method achieves better
performance than the baseline in a metric different from its optimization
objective. This indicates that the higher weight imposed on samples with high
loss improves the overall generalization, which is aligned with the conjecture
proposed in Sec.~\ref{sec:introduction}.

\begin{table}[t]
  \centering
  \caption{Test mean losses at the iteration in which the validation set had
    smallest mean loss. Mean and standard deviations over 50 runs are shown. All
    differences are statistically significant ($p \ll 0.001$).}
  \label{tab:test_losses}
  \begin{tabular}{|c|c|c|}
    \hline
    Corruption level & Mean loss & Hypervolume
    \\
    \hline
    0.0 & 53.604 (0.022) & 53.523 (0.021)
    \\
    0.1 & 57.269 (0.056) & 57.101 (0.057)
    \\
    0.2 & 64.227 (0.131) & 63.723 (0.137)
    \\
    0.3 & 71.385 (0.207) & 70.556 (0.181)
    \\
    0.4 & 78.160 (0.358) & 77.136 (0.389)
    \\
    \hline
  \end{tabular}
\end{table}

Another effect noted in Fig.~\ref{fig:results_mean} is that the noise level
influences the overall behavior of the difference between the methods. For $p
\le 0.3$, the hypervolume method always achieves better results faster than the
baseline and most of the times for $p = 0.4$. As the number of iterations
increase, the difference between the two methods decreases for $p \le 0.3$,
and increases for $p=0.4$. Notice also that the difference in performance is
more favorable to the hypervolume method when the noise level gets higher.
If this occurred only for the training set, it would
indicate an overfitting problem. But since it occurs on all sets, it means
that the hypervolume method is able to cope better with higher noise levels,
increasing the generalization.

\begin{figure}[h]
  \centering
  \begin{subfigure}{0.45\linewidth}
    \psfrag{iterations}[t][c]{\scriptsize Epoch}
    \psfrag{h0-h1}[c][c]{\scriptsize Difference}
    \includegraphics[width=\linewidth]{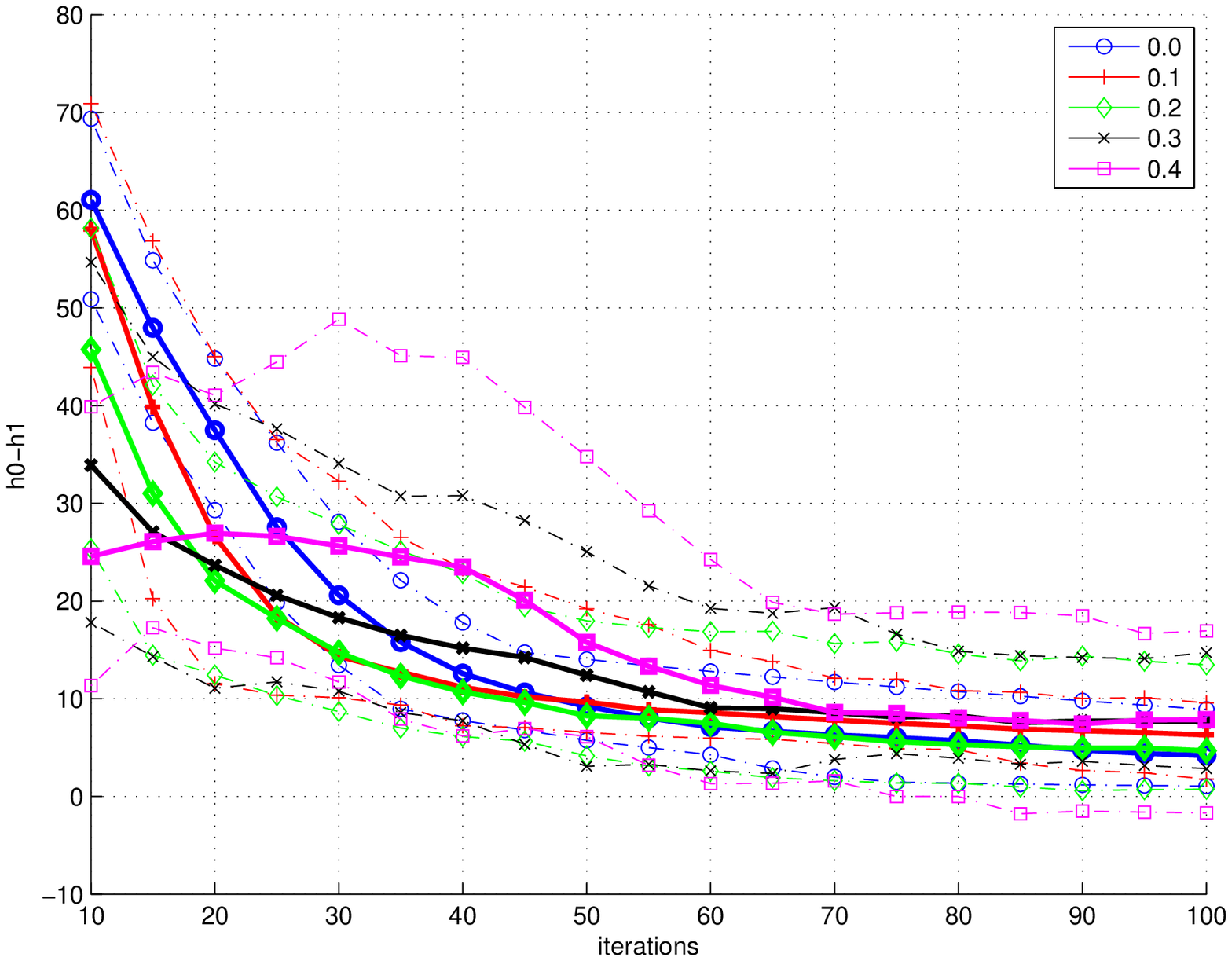}
    \caption{Training}
  \end{subfigure}
  \begin{subfigure}{0.45\linewidth}
    \psfrag{iterations}[t][c]{\scriptsize Epoch}
    \psfrag{h0-h1}[c][c]{\scriptsize Difference}
    \includegraphics[width=\linewidth]{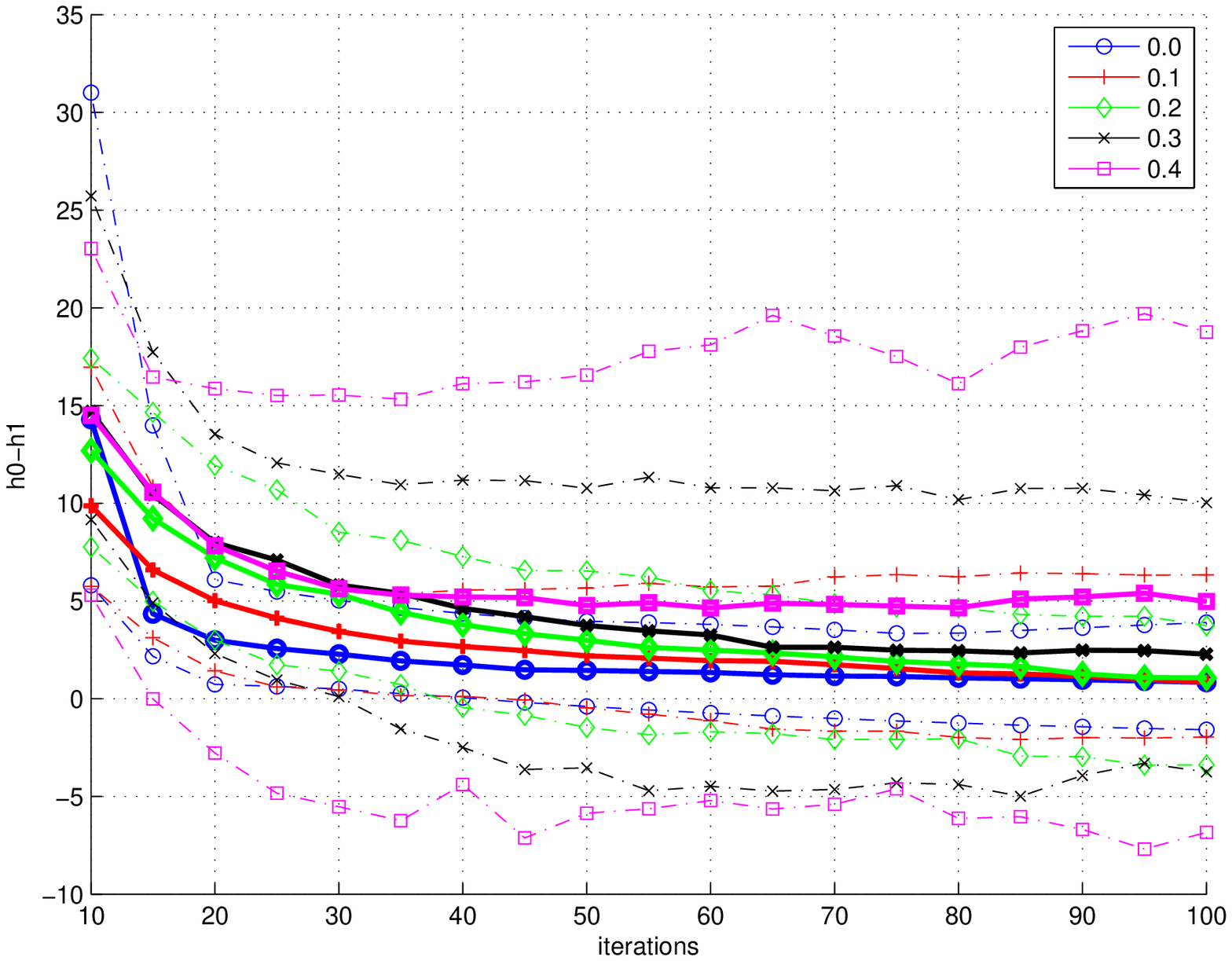}
    \caption{Validation}
  \end{subfigure}
  \\
  \begin{subfigure}{0.45\linewidth}
    \psfrag{iterations}[t][c]{\scriptsize Epoch}
    \psfrag{h0-h1}[c][c]{\scriptsize Difference}
    \includegraphics[width=\linewidth]{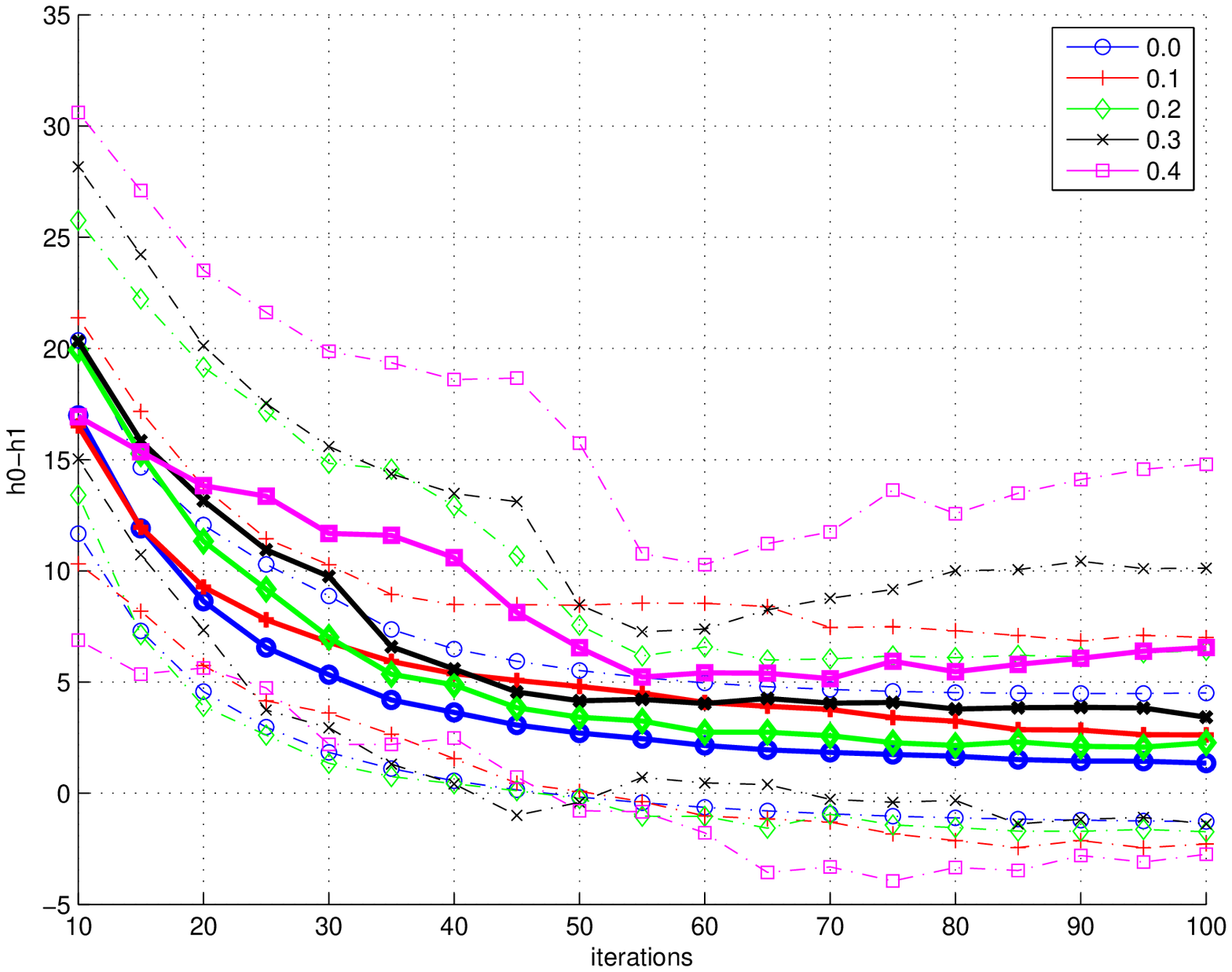}
    \caption{Test}
  \end{subfigure}
  \caption{Difference of the maximum loss between mean loss minimization and
    hypervolume maximization (values above zero mean improvement) for different
    corruption probabilities $p$. The median over the runs is shown in solid
  line, while the upper and lower bounds are shown in dashed lines.}
  \label{fig:results_max}
\end{figure}

Figure~\ref{fig:results_max} shows the maximum loss over the different parts of
the data set as the learning advances. For all parts of the data set, in some
cases the worst-case loss is increased by the hypervolume method, but the median
shows that improvement happens in most cases, with possibly large improvements,
as shown by the upper bounds. The increase in the worst-case loss in the
training data set is explained by the fact that this performance is measured on
the noiseless data, while the training happens on the noisy data.

Table~\ref{tab:test_losses} shows the resulting mean losses for both methods on
the test set. The point of evaluation was chosen as the one where the noiseless
validation set presented the smallest mean loss. Clearly the hypervolume method
provides better results than the commonly used mean loss for all corruption
levels, and the differences are significant at less than 0.1\% using the paired
t-test.
\section{Conclusion}
\label{sec:conclusion}
This paper presented the problem of minimizing the loss function over a data set
from a multi-objective optimization perspective. It discussed the issues of
using linear combination of loss functions as the target of minimization in
machine learning, proposing the use of another metric, the hypervolume
indicator, as the target to be optimized. This metric preserves optimality
conditions of the losses and may achieve trade-offs between the losses more
aligned to the user's expectations.

It is also shown that the gradient of this metric is equivalent to the gradient
of a weighted mean loss, but without requiring the weights to be determined a
priori. Instead, the weights are determined automatically by the gradient and
possesses a boosting-like behavior, with the losses for the current set of
parameters establishing the weights' values.

Experiments with the MNIST data set and denoising autoencoders show that the new
proposed objective achieves better mean loss over the training, validation, and
test data. This increase in performance occurs even though the performance
indicator is different from the objective being optimized, which could cause the
problem to converge to an optimal solution to the problem but with low
performance in the mean loss metric. This indicates that the hypervolume
maximization is similar to the mean loss minimization, but may be able to
achieve better minima due to the change in the loss landscape, as conjectured in
Sec.~\ref{sec:introduction}.

Results also show that the maximum loss is almost always reduced in the
training, validation, and test sets, with a slight reduction in some cases and
large improvements on others. Moreover, by using the validation set to find the
best set of parameters, the hypervolume method achieved statistically
significant smaller mean loss in the test set, indicating that it was able to
generalize better.

Future research directions involve analyzing the effect of including
regularization terms, which usually are added to the linear combination of loss
functions, as new objectives. The use of this method for larger problems and
models and in other learning settings, such as multi-task learning, where
separating tasks in different objectives is natural, should also be pursued.

\subsubsection*{Acknowledgments}

The authors would like to thanks CNPq for the financial support.

\subsubsection*{References}

\begingroup
\renewcommand{\section}[2]{}
\bibliographystyle{unsrt}
\small{\bibliography{paper}}
\endgroup

\end{document}